\documentclass[conference]{IEEEtran}
\IEEEoverridecommandlockouts
\usepackage{cite}
\usepackage{amsmath,amssymb,amsfonts}

\usepackage{algpseudocode}
\usepackage{graphicx}
\usepackage{url}
\usepackage[hidelinks]{hyperref}
\usepackage{textcomp}
\usepackage{xcolor}
\usepackage[ruled]{algorithm2e}
\usepackage{subcaption}
\usepackage{float}
\usepackage{multirow}
\usepackage{booktabs}
\usepackage{threeparttable}
\usepackage{mathtools}

\def\BibTeX{{\rm B\kern-.05em{\sc i\kern-.025em b}\kern-.08em
    T\kern-.1667em\lower.7ex\hbox{E}\kern-.125emX}}
\begin{document}

\title{Communication-Efficient Federated Knowledge Graph Embedding with Entity-Wise Top-K Sparsification\\
\thanks{Identify applicable funding agency here. If none, delete this.}
}

\author{
\IEEEauthorblockN{1\textsuperscript{st}Xiaoxiong Zhang \IEEEauthorrefmark{1}}
\IEEEauthorblockA{\textit{College of Computing and Data Science} \\
\textit{Nanyang Technological University}\\
Singapore\\
zhan0552@e.ntu.edu.sg}
}

\author{
\IEEEauthorblockN{Xiaoxiong Zhang\IEEEauthorrefmark{1} \IEEEauthorrefmark{3}, Zhiwei Zeng\IEEEauthorrefmark{2}, Xin Zhou\IEEEauthorrefmark{1}, Dusit Niyato\IEEEauthorrefmark{3}, Zhiqi Shen\IEEEauthorrefmark{2}\IEEEauthorrefmark{3}}
\IEEEauthorblockA{\IEEEauthorrefmark{1}Joint NTU-Webank Research Institute on Fintech, Nanyang Technological University, Singapore }
\IEEEauthorblockA{\IEEEauthorrefmark{2}Joint NTU-UBC Research Centre of Excellence in Active Living for the Elderly, Nanyang Technological University, Singapore}

\IEEEauthorblockA{\IEEEauthorrefmark{3}College of Computing and Data Science, Nanyang Technological University, Singapore\\ Email: zhan0552@e.ntu.edu.sg, \{zhiwei.zeng, xin.zhou, dniyato, zqshen\}@ntu.edu.sg}
}

\maketitle
  
\begin{abstract}
Federated Knowledge Graphs Embedding learning (FKGE) encounters challenges in communication efficiency stemming from the considerable size of parameters and extensive communication rounds. However, existing FKGE methods only focus on reducing communication rounds by conducting multiple rounds of local training in each communication round, and ignore reducing the size of parameters transmitted within each communication round. To tackle the problem, we first find that universal reduction in embedding precision across all entities during compression can significantly impede convergence speed, underscoring the importance of maintaining embedding precision. We then propose bidirectional communication-efficient \textbf{FedS} based on Entity-Wise Top-K Sparsification strategy. During upload, 
clients dynamically identify and upload only the Top-K entity embeddings with the greater changes to the server. During download, the server first performs personalized embedding aggregation for each client. It then identifies and transmits the Top-K aggregated embeddings to each client. Besides, an Intermittent Synchronization Mechanism is used by \textbf{FedS} to mitigate negative effect of embedding inconsistency among shared entities of clients caused by heterogeneity of Federated Knowledge Graph. Extensive experiments across three datasets showcase that \textbf{FedS} significantly enhances communication efficiency with negligible (even no) performance degradation. 

% The code is available at: \href{https://anonymous.4open.science/r/FedS-68E4}{\color{blue!50!cyan}\url{https://anonymous.4open.science/r/FedS-68E4}}.

% Apart from \textbf{FedS}, we explored the utilization of Singular Value Decomposition (SVD) for compressing local updates on clients, given the low-rank nature of embedding updates. Specifically, we devised two strategies: (1) direct compression of local updates using SVD after each round of training; (2) incorporation of low-rank SVD training within each round of training, followed by compression of local updates using SVD. Nevertheless, experimental across three datasets show that SVD-based strategies affect embedding performance significantly and is not suitable for communication-efficient FKGE task.  

\end{abstract}


\section{Introduction}\label{introduction}
The Knowledge Graph (KG) organizes real-world knowledge in a structured graph format \cite{shen2022comprehensive}. Federated Knowledge Graph (FKG) compiles multiple KGs from diverse sources, decentralized across clients to ensure data privacy\cite{regulation2016679}. The current approaches to Federated Knowlege Graph Embedding (FKGE) learning is based on the prevailing distributed framework: Federated Learning (FL), which usually includes three steps in a training round: uploading each client's entity embedding to a master server after local training, embedding aggregation of shared entities and downloading the aggregated entity embedding \cite{briggs2020federated,fede,fedec,fedkg,shamsian2021personalized}. In this way, different clients not only improve the quality of learned embeddings by sharing staged training results with each other compared with embedding learned only based on local KG, but also avoid exposing one's own raw data. 

Despite the mentioned benefits, the problem of high communication overhead between clients and the master server has posed a substantial challenge to the federated learning-based FKGE. 
There are extensive and frequent exchanges of parameters (i.e., entity embeddings) between clients and the server, especially when numerous clients engage in the process of collaborative training, coupled with large knowledge graph sizes and high embedding dimension on each client. However, the communication links between the server and clients are usually bandwidth-constrained in various wireless edge network scenarios and participating edge devices may be subject to limited data plans featuring costly network connections \cite{luping2019cmfl}. Hence, the need to transmit a lot of parameters conflicts with limited network bandwidth and the high cost of network connections, impeding the training process and may even render it unfeasible \cite{sattler2019robust}.

The total communication cost is determined by two primary factors: the number of communication rounds and the number of transmitted parameters in each round \cite{jiang2022fwc}. Existing FKGE methods, such as \textbf{FedE} \cite{fede} and \textbf{FedEC} \cite{fedec}, typically involve more local iterations at individual clients and infrequent communication with the server, akin to FedAvg \cite{fedavg}. While this strategic approach has shown partial success in reducing the overall communication cost by minimizing the number of rounds, its effectiveness remains limited, as it does not address another significant factor of communication cost, i.e., the size of transmitted parameters in each round. As a result, the transmission of substantial parameters persists during each communication round. 

This paper aims to further reduce the total communication cost of existing FKGE methods by reducing transmitted parameter size per communication round without significantly compromising performance. Intuitively, compressing entity embeddings to be transmitted by model compression techniques may be potential solution. Considering the popularity and success of Knowledge Distillation (KD) \cite{wu2022communication, zhu2022dualde, wang2021mulde, rao2023parameter, zhu2021student, yang2023knowledge} and Low-Rank Approximation (LRA) \cite{yang2020learning, xu2021singular, tukan2021no} in model compression, we try to integrate them into FKGE, respectively, to conduct compression for embeddings to be transmitted. However, extensive experiments show that both methods significantly slow down convergence and instead increase total communication costs, even with modest compression ratio in each communication round. Based on further analysis, we find that the universal reduction in embedding precision across all entities leads to their ineffectiveness. This is explained in detail in Section \ref{iuepr}.

This prompts us to turn to methods reducing parameter size while preserving entity embedding precision. Considering that Entity-Wise Top-K Sparsification strategy enables us reducing parameters to K entity embeddings and the precision of identified entities is also retained, we further propose a simple yet effective method titled \textbf{Fed}erated Knowledge Graph Embedding with Entity-Wise Top-K \textbf{S}parsification (\textbf{FedS}). \textbf{FedS} can improve bidirectional communication cost and is compatible with many FKGE methods as a constituent. During the upload process, the clients dynamically identity the first K entities with greater changes and only upload those to the server. During the download process, the server first conducts personalized embedding aggregation for each client, and then personalizedly identifies the Top-K aggregated embeddings for each client based on the entity upload frequency rather than based on the quantified changes of embeddings as used in clients, due to the heterogeneity of FKG. Besides, the heterogeneity of FKG can also lead to the embedding inconsistency of shared entities across clients, potentially affecting the embedding learning. We propose the Intermittent Synchronization Mechanism, which involves transmitting all parameters between clients and server at fixed intervals, to mitigate the problem.

The main contributions of this paper are summarized as follows:

\begin{itemize}
\item[$\bullet$] We find that the universal reduction in embedding precision across all entities during compression usually impede convergence speed significantly by extensive experiments, and highlight the importance of maintaining embedding precision. 

\item[$\bullet$] Based on the above finding, we propose \textbf{FedS} which mainly applies a novel Entity-Wise Top-K Embedding Sparsification strategy to reduce FKGE communication overhead. To our best knowledge, the work represents the first attempt to mitigate FKGE communication overhead by reducing transmitted parameters size per communication round. 

\item[$\bullet$] We validate that the proposed \textbf{FedS} can notably enhance communication efficiency with only marginal performance degradation on three datasets with three knowledge graph embedding methods. 

\end{itemize}

\section{RELATED WORK}\label{relatedWork}

\subsection{Federated Knowledge Graph Embedding}

Existing federated learning-based FKGE methods can be broadly classified into two architectural paradigms: the client-server architecture and the peer-to-peer architecture \cite{zhang2024personalized}.

Within the client-server architecture, a central server assumes the role of a master aggregator responsible for aggregating entity embeddings from all clients in each iteration. Subsequently, the aggregated result, termed global entity embeddings, is distributed to clients for their subsequent round of embedding updates. Each client employs a knowledge graph embedding method to conduct local embedding learning, utilizing the global entity embeddings and local triples. The pioneering model in this category is FedE \cite{fede}, wherein the server aggregates clients' entity embeddings through averaging, and each client initializes its local entity embedding using the aggregated result at the onset of each training round. Extending upon FedE, FedEC \cite{fedec} introduces embedding-contrastive learning to align local entity embeddings with aggregated entity embeddings while deviating from the previous round's local entity embedding by a regularization term during client's local training. However, both FedE and FedEC pay less attention to the heterogeneity present in federated knowledge graphs, potentially resulting in a divergence between local optimization and global convergence. To address this, FedLu \cite{zhu2023heterogeneous} proposes mutual knowledge distillation to transfer knowledge from local entity embeddings to global entity embeddings and then reintegrate knowledge from global entity embeddings.

In contrast to the aforementioned methods tailored to scenarios where federated knowledge graphs solely share entities, FedR \cite{fedr} is designed for scenarios where clients share both entities and relations. In this paradigm, clients receive identical embeddings of shared relations from the server and then engage in local embedding learning using local triples and the shared relation embeddings.

The peer-to-peer architecture, characterized by the absence of a centralized coordinator like a server, entails clients collaborating on an equal footing and directly exchanging embedding updates. Notably, FKGE \cite{fedkg} is the singular model operating within this paradigm, addressing scenarios akin to those confronted by FedR. Drawing inspiration from MUSE \cite{lampleword}, FKGE employs a Generative Adversarial Network (GAN) \cite{goodfellow2020generative}  to unify embeddings of shared entities and relations within the knowledge graph. 

Whether based on client-server architecture or peer-to-peer architecture, existing methods aim to improve the quality of learned embeddings with comparatively less emphasis on communication efficiency. They simply involve reducing communication rounds through multiple iterations of local embedding training on clients within each communication round. Notably, however, the reduction of parameters transmitted per communication round has not been achieved, resulting in a sustained high communication load. This study aims to mitigate this problem.

\subsection{Communication Efficiency in Federated Learning}

While FedAVG \cite{fedavg}, known as the basic federated learning algorithm, manages to reduce communication expenses by permitting several local steps, the large size of parameters transmitted in each communication round remains a significant hindrance. Broadly, there are three methods for addressing this issue: structured updates, quantization, and sparsification \cite{zheng2023fedpse}.

The structured updates methodology involves the acquisition of parameter updates from a constrained parameter space characterized by a reduced set of variables. For example, the paper \cite{konecny2016federated} introduces two distinct approaches: low-rank and random mask methods. The former technique mandates that the local update matrix $\mathbf{H}$ possesses a low-rank structure, achieved by expressing $\mathbf{H}$ as the product of two smaller matrices. Conversely, the latter technique constrains the update matrix $\mathbf{H}$ to exhibit sparsity, adhering to a predefined random sparsity pattern. However, a notable limitation of them lies in the necessity of generating fresh predefined patterns for each round and client independently, resulting in diminished adaptability and efficiency. Moreover, they predominantly address the reduction of uplink communication costs and do not effectively mitigate downlink communication costs, as the global model aggregated by the server remains uncompressed.

Unlike structured updates, quantization-based methods \cite{reisizadeh2020fedpaq, shlezinger2020uveqfed, mao2022communication, honig2022dadaquant, bouzinis2023wireless} learn the full local update matrix $\mathbf{H}$ during local training without constraints. They then compress it into a lossy form by mapping high-precision floating-point values to a smaller set of discrete values. This approach has an upper bound on compression ratio due to bit limitations and slows convergence speed, given the nature of trade-off between communication efficiency and model accuracy.

Sparsification-based methods, akin to quantization-based approaches, operate post-local training. They selectively transmit elements with higher magnitudes from parameter matrices, typically through predefined thresholds \cite{strom2015scalable} or sparsity rates \cite{ lin2017deep, aji2017sparse, sattler2019robust, zheng2023fedpse, wu2020fedscr}. Unlike neural network models at which previous sparsification-based methods target, FKGE deals with structural graph data and features by inherent coherence of multiple parameters. In FKGE, $n$ parameters form an embedding and collectively represent the semantic of an entity. Conducting parameter-wise sparsification as previous methods do, can corrupt the semantic integrity of embeddings. Hence, we propose Entity-Wise Top-K Sparsification strategy, which is a significant difference between our method with previous ones.

\section{Methodology}\label{methodology}

In this section, we begin by showing the negative influence of universal embedding precision reduction, which further leads us to the Entity-Wise Top-K Sparsification strategy to reduce FKGE communication costs. We then present an overview of the proposed method, \textbf{FedS}, and subsequently describe its three main components in detail.

\subsection{Influence of Universal Embedding Precision Reduction}\label{iuepr}

In our endeavor to reduce FKGE communication overhead, we initially examine the feasibility of integrating Model Compression (MC) methods into FKGE. Knowledge Distillation (KD) has gained prominence for effectively compressing model parameters with minimal performance decline \cite{wu2022communication}. Besides, Low-Rank Approximation (LRA) such as Singular Value Decomposition (SVD) have also proven effective in MC due to the intrinsic low-rank nature of model parameters \cite{jiang2022fwc}. Given their effectiveness in MC, we particularly investigate their potential for reducing communication cost. 

In terms of introducing KD into FKGE, it involves each client maintaining both low- and high-dimensional embeddings for each entity. Low-dimensional embeddings are used for communication and both embeddings conduct knowledge co-distillation during client update. For LRA, two strategies are employed. The first strategy involves directly applying SVD to the entity embedding update matrix after client local training, retaining only specific higher singular values to obtain smaller matrices for communication. The second strategy enhances the low-rank property of entity update matrices through additional constraints during client local training, followed by SVD-based embedding compression, denoted formally as SVD+. 

However, extensive experiments show that these strategy significantly slow convergence and instead increase total communication costs, even with low compression ratios (25\% for TransE and RotatE in KD; $\textless$ 20\% for TransE and $\textless$ 30\% for RotatE in SVD and SVD+) in each communication round, as Table \ref{moti} shows. 

\begin{table}[h] 
\centering
\tabcolsep=0.12cm
\renewcommand\arraystretch{1.3}
\caption{Comparison of total transmitted parameter size (scaled by those of \textbf{FedE}) across different models\protect\footnotemark, when first reaching 98\% convergence accuracy of \textbf{FedE}.}
\begin{tabular}{llccc} 
\hline 
\multirow{2}{*}{KGE}&  \multirow{2}{*}{Model}  & \multicolumn{3}{c}{Dataset}\\
\cline{3-5}
 & & FB15k-237-R10 &FB15k-237-R5 &FB15k-237-R3\\

\cline{1-5}
\multirow{2}{*}{TransE} & \textbf{FedE}  &1.00x  &1.00x  &1.00x\\
                        & \textbf{FedE-KD}  &1.75x  &2.10x  &2.50x\\
                        & \textbf{FedE-SVD}  &1.39x  &1.44x  &1.33x\\
                        & \textbf{FedE-SVD+} &1.92x  &2.08x  &2.14x\\
                  
\cline{1-5}
\multirow{2}{*}{RotatE}  & \textbf{FedE}  &1.00x  &1.00x  &1.00x\\
                         & \textbf{FedE-KD}  &1.75x  &2.25x  &2.40x\\
                         & \textbf{FedE-SVD}  &1.38x &1.43x &1.28x\\
                         & \textbf{FedE-SVD+} &2.28x &2.23x &1.57x\\          
\hline 
\end{tabular}
\label{moti}
\end{table}
\footnotetext{$\textbf{FedE-KD}$,  $\textbf{FedE-SVD}$ and $\textbf{FedE-SVD+}$ are the models that the KD, SVD and SVD+ strategies are applied to  $\textbf{FedE}$, respectively. The implementation and training details of them are provided in Appendices \ref{kd} and \ref{svd}.}

Analyzing these methods, we find they all reduce communication overhead per round by decreasing embedding precision for all entities. The KD strategy transfers information from high-dimensional to low-dimensional embeddings for all entities, inevitably causing some loss of semantic information. Both the SVD and SVD+ strategies directly lose some information from embeddings. Particularly, SVD+ basically impacts the accurate semantic learning process for all entities by limiting the embedding update matrix in a lower-rank space, given its notable weakness than SVD as Table \ref{moti} shows. Applying these strategies to all entities results in embedding precision loss across all entities. Moreover, aggregating these inaccurate embeddings in the server exacerbates bias in the obtained embeddings. We believe that it is the reason for their ineffectiveness. This further leads us to strategy preserving embedding precision of certain entities while the size of transmitted parameters decreases.  

Naturally, we consider the Entity-Wise Top-K Sparsification strategy, which addresses the need to preserve embedding precision for specific entities by identifying and transmitting original entity embeddings containing more valuable information. Hence, we propose the method: \textbf{FedS}, based on the Entity-Wise Top-K Sparsification strategy. 

\subsection{Overview of \textbf{FedS}}

For \textbf{FedS}, the process remains consistent across different communication rounds and different clients. We take client $c$ at round $t$ as an example to explain the proposed method. By default, the term `` entities of client $c$ '', in the following parts, specifically refers to client $c$'s entities shared with at least one other client, as exclusive entities do not need to be involved in the communication. The size of this set is denoted as $\mathrm{N}_c$.

\begin{figure}
\centerline{\includegraphics[height=2.5 in]{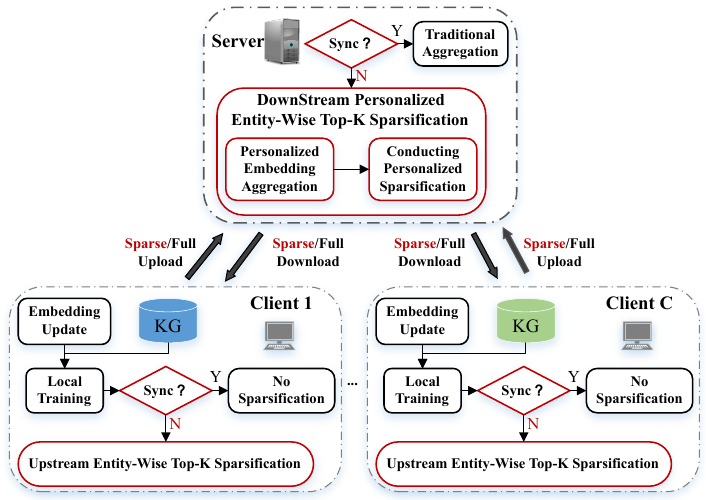}}
\caption{Overall procedure of \textbf{FedS} at each communication round.} \label{FedS}
\vspace{-1.8 em}
\end{figure}

\textbf{FedS}, as a constituent, is integrated into existing FKGE methods to reduce the communication cost. Figure \ref{FedS} illustrates how \textbf{FedS} operates during each communication round. The functions of \textbf{FedS} are highlighted in the red frame. \textbf{FedS} mainly includes three components: Upstream Entity-Wise Top-K Sparsification, Downstream Personalized Entity-Wise Top-K Sparsification and Intermittent Synchronization Mechanism. 

After local training, clients assess if the current round meets the criteria for entity embedding synchronization, governed by the Intermittent Synchronization Mechanism. If synchronization is required, the clients apply Upstream Entity-Wise Top-K Sparsification, selecting and transmitting only the top-K entity embeddings with greater changes to the server. Otherwise, all entity embeddings are transmitted.

Upon receiving the uploads, the server also determines if embedding synchronization is satisfied. If so, the server performs Downstream Personalized Entity-Wise Top-K Sparsification. This involves personalized embedding aggregation and personalized sparsification for each client, followed by sending the selected embeddings back to clients. Otherwise, the server aggregates and transmits all entity embeddings to clients, following the standard procedure of the chosen FKGE method.

Detailed explanations of these three components are provided in the following subsections.

% Due to the heterogeneity of FKG, the changes-based Top-K strategy used in clients is impractical for server. Instead, the server applies Downstream Personalized Entity-Wise Top-K Sparsification which is based on the upload frequency of entities, identifying the Top-K aggregated entity embeddings for each client in a personalized manner. The Intermittent Synchronization Mechanism aims to synchronize entity embeddings across clients at fixed intervals to mitigate the negative impact of accumulated embedding inconsistencies, which are also caused by the heterogeneity of FKG. Detailed explanations of these three components are provided in the following subsections.

\subsection{Upstream Entity-Wise Top-K Sparsification}

This module, adopted in clients, only aims to perform Entity-Wise embedding sparsification after local training of clients in each round and is not related to the local training process itself. 
% Necessary information about local training is provided in subsection \ref{downstream} for better understanding.

Previous Top-K strategies in Federated Learning, usually conducts sparsification in parameter-wise manner (i.e., each parameter undergoes Top-K selection independently). However, FKGE handles structural graph data and features by the inherent coherence of multiple parameters (i.e., $n$ parameters form an embedding and collectively represent the semantic of
an entity). Instead of performing sparsification in the parameter-wise manner, we propose conducting sparsification in entity-wise manner, which asks each entity embedding to be subject to Top-K selection, and hence retain the semantic integrity of embeddings. 

Specifically, each client keeps the history upload embeddings $\mathbf{E}^h_c \in \mathbb{R}^{\mathrm{N}_c \times m}$ ($m$ is the embedding dimension) for its entities, representing the latest embeddings sent to the server for each entity. They are initialized as the same values as the local entity embeddings at round 0 ($\mathbf{E}^0_c \in \mathbb{R}^{\mathrm{N}_c \times m}$). We identify Top-K entity embeddings by quantifying the embedding changes for each entity, through Cosine Similarity between the entity's current embedding and the latest embedding sent to the server. A higher Cosine Similarity corresponds to smaller change. Formally, the changes of entity embeddings for client $c$ at round $t$, denoted as $\mathbf{M}^{t}_{c}$, is expressed as:
\begin{equation}
    \mathbf{M}^{t}_{c} = \mathbf{I} - \cos{(\mathbf{E}^t_c, \mathbf{E}^h_c)}
\end{equation}
where $\mathbf{I} \in \mathbb{R}^{\mathrm{N}_c}$ is a unit vector;  $\mathbf{E}^t_c \in \mathbb{R}^{\mathrm{N}_c \times m}$ is the entity embeddings of client $c$ at round $t$; $\mathbf{E}^h_c \in \mathbb{R}^{\mathrm{N}_c \times m}$ is client $c$'s history upload embeddings.

With the quantified changes of embeddings for all entities, the client $c$ selects the first $\mathrm{K}$ entity embeddings with more notable changes and sends them to the server. The $\mathrm{K}$ is defined as follows:
\begin{equation}\label{eq_k}
    \mathrm{K} = \mathrm{N}_c \times p
\end{equation}
where $p$ is the sparsity ratio. 

Together with selected entity embeddings, the 0-1 sign vector $\mathbf{S}^t_c \in \mathbb{R}^{\mathrm{N}_c}$, representing whether entities of client $c$ is selected in current round $t$, is also sent to the server.

At last, updating corresponding embeddings in $\mathbf{E}^h_c$ for selected Top-K entities as their current embeddings.

\subsection{Downstream Personalized Entity-Wise Top-K Sparsification}\label{downstream}

The objective of Downstream Personalized Entity-Wise Top-K Sparsification entails the aggregation of entity embeddings and the selection of the Top-K entity embeddings tailored to each client.

Due to the inherent data heterogeneity among clients, considerable differences exist among the Top-K entities uploaded by different clients during a communication round. It is very likely that the embedding of a specific entity (e.g., $e$) may not be transmitted by client $c$ while being transmitted by others. Consequently, employing the Cosine Similarity-based Top-K strategy used by clients to measure the change of the aggregated embedding of $e$ to the embedding of $e$ of client $c$ becomes impractical. Even if the server keeps the uploaded history embeddings for clients, the practical issue also arises. For different entities of client $c$, the server usually retains history embeddings from different rounds, even with considerable discrepancies in rounds. Consequently, the disparity between the current round of embeddings and those kept by the server can be substantial for different entities of $c$. This further results in notable bias in quantified changes of the aggregated relative to the history embeddings across different entities.  

We notice that, for different entities of client $c$, the frequency of other clients uploading those entities in a communication round can present notable difference. Intuitively, as more clients upload embeddings for an entity, the aggregated embedding for the entity holds greater information and significance. However, the ranking of a entity's upload frequency may vary among different clients. Hence, we propose Personalized Entity-Wise Top-K Sparsification, which, in client-specific manner, ranks aggregated embeddings based on entities' upload frequency and selects Top-K ones. Formally, the entity upload frequency is denoted as \textbf{priority weight}.

Specifically, the server first conducts personalized entity embedding aggregation for each client (e.g. $c$). For every entity $e$ of client $c$, the server aggregates the embeddings of the respective entity 
$e$'s embedding receiving from other clients. It is emphasized that, different from the aggregation way used by previous FKGE method, the aggregation process for client $c$ is not involved in $c$'s entity embeddings, since not all of $c$'s entity embeddings are uploaded due to the Top-K strategy in clients. Formally, the aggregated entity embedding for entity $e$ of client $c$ at round $t$, denoted as $\mathbf{A}^t_{c_e}$, is expressed as:
\begin{equation}\label{equ_A}
    \mathbf{A}^t_{c_e} = \sum_{i\in C^t_{c_e}} \mathbf{E}^t_{i_e}
\end{equation}
where $C^t_{c_e}$ denotes the set of clients that transmit the embedding of entity $e$ in round $t$, while $\mathbf{E}^t_{i_e}$ represents the entity $e$'s embedding of client $i$ in round $t$.

Subsequently, the server proceeds to sparsity the aggregated embeddings for client $c$ by selecting the first $\mathrm{K}$ entity embeddings. The priority of each entity (e.g., $e$) is determined by the number of clients from which its aggregagted embedding originates (i.e., the size of $C^t_{c_e}$). The value of $\mathrm{K}$ is determined by Eq. \ref{eq_k}.  In cases where the number of available aggregated entity embeddings is less than $\mathrm{K}$, the server transmits all available aggregated entity embeddings to client $c$. In scenarios where multiple entities of equal priority compete to satisfy the requirement of $\mathrm{K}$, a random strategy is employed.

Finally, the server sends to clients (e.g. $c$) the aggregated entity embedding, \textbf{priority weight} vector $\mathbf{P}^t_c \in \mathbb{R}^\mathrm{K}$ and 0-1 sign vector $\mathbf{S}^t_c \in \mathbb{R}^{\mathrm{N}_c}$, to empower the next round of local training of clients. The 0-1 sign vector $\mathbf{S}^t_c \in \mathbb{R}^{\mathrm{N}_c}$ represents whether the corresponding aggregated entity embedding is sent. Specifically, each client first combines its local embeddings with the aggregated embeddings from the server for embedding update. Formally, for each entity $e$ whose embedding needs to be updated according to $\mathbf{S}^t_c$, its updated embedding is as follows:
\begin{equation}
    \mathbf{E}^{t+1}_{c_e} = \frac{1}{1+\mathbf{P}^t_{c_e}}\sum (\mathbf{A}^t_{c_e} + \mathbf{E}^t_{c_e})
\end{equation}
where $\mathbf{P}^t_{c_e}$ is the entry of $\mathbf{P}^t_c$ corresponding to entity $e$, the value of which is the size of $C^t_{c_e}$; $\mathbf{E}^t_{c_e}$ is embedding of entity $e$ of client $c$ at round $t$; $\mathbf{A}^t_{c_e}$ is explained in Eq. \ref{equ_A}. 

After that, each client conducts next round of local training with chosen FKGE methods.  

\subsection{Intermittent Synchronization Mechanism}

As discussed in Section \ref{downstream}, the variance in data heterogeneity contributes to differences among the Top-K entities uploaded by various clients and also influences the prioritization of sparsification for identical entities across different clients. Consequently, it is highly probable that identical entities across different clients receive disparate updates during a communication round. Consider a scenario where there are three clients, all possessing entity $e$. In communication round $t$, only client $a$ transmits the embedding of $e$ to the server. If both client $b$ and $c$ receive the aggregated embedding of $e$ (i.e., client $c$'s $e$'s embedding) from the server, the updated embedding of $e$ for clients $b$ and $c$ is the average between their local embedding and the embedding from the server. The embedding of $e$ for client $a$ remains unchanged. 
This scenario illustrates the discrepancy in updates received by entity $e$ across the three clients. 

With the progression of communication rounds, the inconsistencies in entity embeddings across clients may intensify, potentially affecting the training process. To alleviate the cumulative effects of these inconsistencies in entity embeddings updates across clients, we propose a simple yet efficient Intermittent Synchronization Mechanism. This mechanism entails clients and the server exchanging all parameters at fixed intervals of communication rounds, thereby facilitating the synchronization of embeddings for identical entities across all clients. It functions by having the clients and server check if the difference between the current round and the last synchronization round matches a predefined interval, before deciding whether to conduct sparsification.

\subsection{Analysis of Communication Efficiency}

We define the period between one synchronization stage and the next (inclusive) as a \textbf{cycle}, and proceed to analyze the communication efficiency of FedS within this cycle. We simply use client $c$ as an illustrative example considering the commonality across clients. We designate the sparsity ratio as $p$, the embedding dimension as $\mathrm{D}$ and the synchronization interval is $s$ signifying there are $s$ communication rounds between two consecutive synchronization operations (exclusive). Besides, it is presumed that client $c$ owns $\mathrm{N}_c$ shared entities with the other clients, which are necessary to transmit corresponding embeddings to the server.

From the sparsification process of clients and the server, it is found that client $c$ transmits $\mathrm{K}$ entity embeddings (with a parameter count of $\mathrm{N}_c \times \mathrm{D} \times p$), along with a 0-1 sign vector $\mathbf{S}^t_c \in \mathbb{R}^{\mathrm{N}_c}$, a process mirrored by the server. Moreover, the server transmits an entity priority vector $\mathbf{P}^t_c \in \mathbb{R}^\mathrm{K}$ ($\mathrm{K} = \mathrm{N}_c \times p$). In the synchronization process, both the server and client $c$ exchange all parameters with each other, amounting to $\mathrm{N}_c \times \mathrm{D}$. In contrast, traditional FKGE methods always exchange all parameters between the server and clients (i.e., $\mathrm{N}_c \times \mathrm{D}$) in each communication round. Formally, the ratio $\mathrm{R}^p_c$ of parameters transmitted by \textbf{FedS} compared to traditional methods is expressed as: 
\begin{equation}\label{equ_ace}
\begin{aligned}
    \mathrm{R}^p_c &= \frac{2(\mathrm{N}_c \times \mathrm{D} \times p \times s +  \mathrm{N}_c \times \mathrm{D} ) + 2 \mathrm{N}_c \times s + \mathrm{N}_c \times p \times s}{2 \mathrm{N}_c \times \mathrm{D} \times(s+1)}\\
    &= \frac{ p \times s +  1  + \frac{(2+p)\times s}{2 \mathrm{D}}}{s+1}\\
\end{aligned}
\end{equation}

Notably, the computed $\mathrm{R}^p_c$ represents a worst-case scenario, with the actual ratio potentially lower. Due to employing the same sparsity ratio across clients, a client (e.g., $c$) with a greater number of entities may be unable to obtain the required $\mathrm{K}$ entity embeddings from the server when other clients lack sufficient entities. It may still occur even if other clients possess a greater number of entities than client $c$ but fail to upload sufficient embeddings for entities owned by $c$. Besides, each element of sign vector may utilize a 1-bit data type. However, both elements of sign vector and entity embedding use the same data type (usually a 32-bit float) in the formula.  
% Besides, we assume that the entries in the sign vector adhere to the same data format as the entity embeddings in Equation \ref{equ_ace}. However, the entries in the sign vector can be represented as a single bit, as opposed to the floating-point data typically used (usually 32 bits) in the entries of entity embeddings.

\section{EXPERIMENTS}\label{experiments}
In this section, we apply the proposed \textbf{FedS} to the pioneering FKGE model \textbf{FedE} and assess the improvement of communication efficiency on real-world datasets. 

\subsection{Dataset}
We conduct experiments using three datasets specifically designed for FKGE task: FB15k-237-R10, FB15k-237-R5, and FB15k-237-R3. They stem from the widely utilized KG embedding dataset FB15k-237 and are created by partitioning relations evenly and then distributing corresponding triples into ten, five, and three clients, respectively.
All of them adhere to the same ratio for dividing training, validation, and testing triples: 0.8/0.1/0.1.

\subsection{Experimental Setup}\label{es}

In this study, we opt for the pioneering FKGE model \textbf{FedE} as the foundation and incorporate \textbf{FedS} into it to assess the enhancement in communication efficiency. \textbf{FedE} learns a unified global embedding for all clients, while the \textbf{FedS} component leads to personalized embeddings for individual clients. Owing to the intrinsic benefits of personalized models over global models, we initially enhance the original \textbf{FedE} framework to produce a personalized edition (i.e., \textbf{FedEP}). This is achieved by incorporating local embeddings to assess its link prediction performance on validation and testing sets throughout the training process. Subsequently, we employ \textbf{FedEP} as the baseline for a fair comparison. Additionally, we add the baseline scenario \textbf{FedEPL}, wherein we directly \textbf{L}ower the embedding dimension of \textbf{FedEP} to ensure that the size of parameters transmitted in a cycle equals that of \textbf{FedS}, while keeping other parts of \textbf{FedEP} unchanged. \textbf{FedS} should have less communication cost than \textbf{FedEP} when achieving the same accuracy, to further support its effectiveness. Additionally, we adopt the baseline scenario denoted as \textbf{Single}, wherein KGE is executed individually for each client solely utilizing its local data. During local training of clients, we adhere to the convention established by prior FKGE methods, selecting three representative KGE methods: TransE \cite{transe}, RotatE \cite{rotate} and ComplEx \cite{complex}.   

We focus on predicting the tail (head) entity when provided with the head (correspondingly, tail) entity and relation. We assess the prediction accuracy  with two metrics:  Mean Reciprocal Rank (MRR) and Hits@10. The overall metric value is derived by aggregating all clients' values through weighted average, with weights being the proportions of the triple size. Both metrics represent the outcomes attained when the model \textbf{C}onver\textbf{G}es. We sometimes use MRR@CG to denote MRR. 
 
For the assessment of communication efficiency, we introduce three metrics: P@CG, P@99 (or P@98) and R@CG. P@CG represents the total transmitted parameters when model converges. P@99(or P@98) quantifies the ratio of the transmitted parameters between a model and \textbf{FedEP}, regarding the first attainment of 99\% (98\%) accuracy in \textbf{FedEP}'s MRR@CG. For easier comparison, the three metrics of a model are scaled by corresponding metrics of the baseline \textbf{FedEP}. The lower these three metrics are, the more effective the model is. R@CG means the communication rounds when a model converges.

We assume that all clients participate in each communication round. The parameter settings across all experiments are as follows: batch size, local epochs, and embedding dimension for client training are 512, 3, and 256, respectively. Initialization parameters $\gamma$ and $\epsilon$ are 8 and 2, respectively. An early stopping mechanism with a patience parameter setting as 3 is implemented, signifying that training ceases after three consecutive declines in MRR of the validation set. The synchronization interval $s$ is 4. In the \textbf{Single} setting, personalized embeddings are evaluated on validation sets every 10 rounds, while for other models, this occurs every 5 epochs. Adam \cite{adam} optimizer with a learning rate of 0.0001 is used. Self-adversarial negative sampling is applied to TransE and RotatE models with a temperature of 1. The sparsity ratio $p$ is set as 0.7 for experiments conducted on FB15k-237-R5, employing ComplEx as the KGE method. In all other instances, the sparsity ratio $p$ is set as 0.4. The embedding dimensions for FedEPL for the two cases are 196 and 135, respectively. The computation method is detailed in Appendix \ref{ced} for reference. 

\subsection{Quantitative Analysis}
This section assesses the efficacy of \textbf{FedS} in enhancing communication efficiency by addressing three critical inquiries:
\begin{itemize}
\item[$\bullet$] Can \textbf{FedS} achieve the same high prediction accuracy (i.e., 98\% and 99\% MRR@CG of \textbf{FedEP}) with fewer transmitted parameters than \textbf{FedEP}? 

\item[$\bullet$] Can \textbf{FedS} achieve comparable prediction accuracy to \textbf{FedEP} after converges, with fewer transmitted parameters than \textbf{FedEP}?

\item[$\bullet$] Can FedS achieve higher prediction accuracy than \textbf{FedEPL} with the same size of transmitted parameters?

\end{itemize}

To answer the first two questions, experiments are conducted on three datasets for \textbf{FedS} and \textbf{FedEP}, and the experiment results are reported in Table \ref{main_result1} and \ref{main_result2}. 
% \begin{table}[h]
% \centering
% \tabcolsep=0.18cm
% \renewcommand\arraystretch{1.1}
% \begin{tabular}{llll lll} 
% \hline 
% \multirow{2}{*}{KGE}&  \multirow{2}{*}{Setting}  & \multicolumn{5}{c}{FB15k-237-R3}\\
% \cline{3-7}
%  & &MRR &Hits@10& P@CG & P@99 & P@98\\
% \cline{1-7}
% \multirow{4}{*}{TransE}       
%                   & Single &0.3229  &0.5608  &-  &- &-\\
%                   & FedEP &\textbf{0.3612} &\textbf{0.6070}  &1  &1  &1\\
%                   & FedS &0.3588 &0.6039 &47.82\%  &81.47\%  &69.83\%\\
                  
% \cline{1-7}
% \multirow{4}{*}{RotatE}       
%                   & Single &0.3409  &0.5643  &-  &- &-\\
%                   & FedEP &\textbf{0.3702} &\textbf{0.6129}  &1  &1  &1\\
%                   & FedS &0.3686 &\textbf{0.6129} &66.16\%  &85.11\%  &73.34\% \\
                  
% \cline{1-7}
% \multirow{4}{*}{ComplEx}       
%                   & Single &0.3029  &0.4724  &-  &- &-\\
%                   & FedEP &\textbf{0.3198} &\textbf{0.5346} &1  &1  &1\\
%                   & FedS &0.3170 &0.5252 &52.38\%  &62.85\%  &52.38\%\\
                  
% \hline 
% \end{tabular}
% \caption{Communication effectiveness of the proposed method FedS compared with the baseline: FedE on FB15k-237-R3. The symbol ``-'' denotes unavailability.}
% \label{main_result2}
% \end{table}

Through a comparative analysis of the performances of \textbf{FedS} and \textbf{FedEP} based on metrics P@99 and P@98, it is evident that \textbf{FedS} can achieve high accuracy with fewer parameters than \textbf{FedEP}. Specifically, when utilizing TransE and RotatE as KGE methods on FB15k-237-R10, \textbf{FedS} requires only 44.11\% and 47.14\%, respectively, of the parameters necessary for \textbf{FedEP} to attain 99\% accuracy of MRR@CG. This results in a saving of about 56\% and 53\% of parameters, respectively. Similarly, the corresponding reduction when achieving 98\% accuracy of MRR@CG of \textbf{FedEP} is more than 54\% and 51\%. Due to the inherent inferiority of \textbf{FedEP} compared to \textbf{Single} when utilizing ComplEx as the KGE method, we omit the results of \textbf{FedS}. When employing TransE, RotatE, and ComplEx as KGE methods on FB15k-237-R5, and aiming to achieve 99\% accuracy of MRR@CG of \textbf{FedEP}, the associated reductions in parameter quantity are more than 55\%, 33\%, and 23\%, respectively. Similarly, the corresponding reductions when aiming for 98\% accuracy of MRR@CG of \textbf{FedEP} are about 53\%, 41\%, and 14\%. On FB15k-237-R3, parameters are reduced by about 19\%, 15\%, and 37\% for TransE, RotatE, and ComplEx, respectively, when achieving 99\% accuracy of MRR@CG of \textbf{FedEP}. When aiming for 98\% accuracy of MRR@CG of \textbf{FedEP}, the associated reductions are more than 30\%, 26\%, and 47\% for TransE, RotatE, and ComplEx, respectively. Based on these data, it can be further found that the enhancement in communication efficiency of \textbf{FedS} is more pronounced when the dataset comprises more clients. Moreover, in most cases, the improvement is notably greater when usinging TransE or RotatE compared with ComplEx.

\begin{table}[h]
\centering
\tabcolsep=0.08cm
\vspace{1.1em}
\renewcommand\arraystretch{1.4}
\caption{Comparison of predication accuracy between \textbf{FedS} and \textbf{FedEP} on FB15k-237-R10, FB15k-237-R5 and FB15k-237-R3. We do not show the result of \textbf{FedS} due to \textbf{FedEP}'s inferior MRR compared with \textbf{Single}. The bold denotes the best result.}
\begin{tabular}{lll lllll ll} 
\hline 
\multirow{2}{*}{KGE}&  \multirow{2}{*}{Setting}  & \multicolumn{2}{c}{FB15k-237-R10}&  &\multicolumn{2}{c}{FB15k-237-R5}& &\multicolumn{2}{c}{FB15k-237-R3}\\
\cline{3-4}
\cline{6-7}
\cline{9-10}
&&MRR &Hits@10 && MRR &Hits@10 & &MRR &Hits@10 \\
\cline{1-10}
\multirow{4}{*}{TransE}       
                  & Single &0.2869 &0.5244  & &0.3014 &0.5335 & &0.3229 & 0.5608  \\
                  \cline{2-10}
                  & FedEP &0.3517 &0.6104   & &\textbf{0.3626} &\textbf{0.6102}& & \textbf{0.3612} &\textbf{0.6070}  \\
                  & FedS &\textbf{0.3541} &\textbf{0.6121}  & &0.3618 &0.6098  & & 0.3588 &0.6039\\
                  
\cline{1-10}
\multirow{4}{*}{RotatE}       
                  & Single &0.3038 &0.5095  & &0.3193 &0.5335  & & 0.3409  &0.5643\\
                  \cline{2-10}
                  & FedEP &0.3657 &0.6184  & &\textbf{0.3723} &0.6184 & & \textbf{0.3702} &\textbf{0.6129} \\
                  & FedS &\textbf{0.3676} &\textbf{0.6200} &&0.3718 &\textbf{0.6193} && 0.3686 &\textbf{0.6129}\\
                  
\cline{1-10}
\multirow{4}{*}{ComplEx}       
                  & Single &\textbf{0.3002} &0.4713  &&0.3013 &0.4645 & & 0.3029  &0.4724\\
                  \cline{2-10}
                  & FedEP &0.2986 &\textbf{0.5297}  &&\textbf{0.3056} &\textbf{0.5205} & & \textbf{0.3198} &\textbf{0.5346} \\
                  & FedS &-  &-  &&0.3029 &0.5189  && 0.3170 &0.5252\\
                  
\hline 
\end{tabular}
\label{main_result1}
\end{table}

\begin{table}[h]
\centering
\tabcolsep=0.08cm
\vspace{1.1em}
\renewcommand\arraystretch{1.3}
\caption{Comparison of communication overhead between \textbf{FedS} and \textbf{FedEP} on FB15k-237-R10, FB15k-237-R5 and FB15k-237-R3, when attaining certain prediction accuracy. We do not show the result of \textbf{FedS} due to \textbf{FedEP}'s inferior MRR compared with \textbf{Single}.}
\begin{tabular}{lll lllll ll} 
\hline 
\multirow{2}{*}{KGE}&  \multirow{2}{*}{Metric}  & \multicolumn{2}{c}{FB15k-237-R10}&  &\multicolumn{2}{c}{FB15k-237-R5}& &\multicolumn{2}{c}{FB15k-237-R3}\\
\cline{3-4}
\cline{6-7}
\cline{9-10}
&&FedEP &FedS && FedEP &FedS & &FedEP &FedS \\
\cline{1-10}
\multirow{4}{*}{TransE}       
                  & P@CG &1.00x &0.5238x  & &1.00x &0.4400x & &1.00x  &0.4782x  \\
                  & P@99 &1.00x &0.4411x  & &1.00x &0.4489x & &1.00x  &0.8147x  \\
                  & P@98 &1.00x &0.4539x  & &1.00x &0.4714x & &1.00x  &0.6983x  \\
                  
\cline{1-10}
\multirow{4}{*}{RotatE}       
                  & P@CG &1.00x &0.5836x  & &1.00x &0.5396x & &1.00x  &0.6616x  \\
                  & P@99 &1.00x &0.4714x  & &1.00x &0.6666x & &1.00x  &0.8511x  \\
                  & P@98 &1.00x &0.4888x  & &1.00x &0.5892x & &1.00x  &0.7334x  \\
                  
\cline{1-10}
\multirow{4}{*}{ComplEx}       
                  & P@CG &1.00x &-         & &1.00x &0.7642x & &1.00x  &0.5238x  \\
                  & P@99 &1.00x &-         & &1.00x &0.7642x & &1.00x  &0.6285x  \\
                  & P@98 &1.00x &-         & &1.00x &0.8600x & &1.00x  &0.5238x  \\
                  
\hline 
\end{tabular}
\label{main_result2}
\end{table}

Through a comparative analysis of the performances of \textbf{FedS} and \textbf{FedEP} utilizing metrics such as MRR, Hits@10, and P@CG, it is noted that there is no significant decline in performance (interestingly, a slight improvement is observed on FB15k-237-R10) when \textbf{FedS} converges compared to \textbf{FedEP}, while there is a notable reduction in the quantity of parameters transmitted by \textbf{FedS}. In particular, when utilizing TransE, RotatE, and ComplEx as KGE methods on FB15k-237-R5, the MRR of \textbf{FedS} at convergence achieves 99.78\%, 99.87\%, and 99.12\% of that of \textbf{FedEP}, respectively. The reduction in transmitted parameter quantity exceeds 56\%, 46\%, and 23\%, respectively. Similarly, on FB15k-237-R3, \textbf{FedS} achieves MRR@CG of 99.34\%, 99.57\%, and 99.12\% of that of \textbf{FedEP}, with parameter quantity reductions exceeding 52\%, 33\%, and 47\%, respectively. Surprisely, \textbf{FedS} achieve higher MRR than \textbf{FedEP} by 0.24\% and 0.19\% when TransE and RotatE are used, respectively, with parameter quantity reductions exceeding 47\% and 41\%. The metric Hits@10 witnesses a similar trend across all instances. One potential explanation for this increase is that the aggregated embedding from the server may not consistently provide useful information due to data heterogeneity. In \textbf{FedS}, clients only update portions of their entity embeddings with those from the server, thereby mitigating information disturbance.

To answer the third question, experiments are conducted on three datasets for \textbf{FedS} and \textbf{FedEPL}, and the experiment results are reported in Table \ref{main_result3}. It indicates that \textbf{FedS} achieves higher accuracy than \textbf{FedEPL} while transmitting fewer parameters overall, which further underscores the superiority of \textbf{FedS} over the method of directly reducing the embedding dimension of the base model to the same level as \textbf{FedS}. 

In particular, when employing TransE as the KGE method on FB15k-237-R10, FB15k-237-R5 and FB15k-237-R3, \textbf{FedS} demonstrates superior performance to \textbf{FedEPL} in terms of MRR by 1.2\%, 0.94\%, and 0.87\%, respectively. Moreover, the reduction in transmitted parameters relative to \textbf{FedEPL} corresponds to 56.58\%, 65\%, and 43.24\%, respectively, reflecting substantial savings (Note that the ratio of transmitted parameter quantity between \textbf{FedS} and \textbf{FedEPL} equals the ratio of their communication rounds, as indicated in Section \ref{es}). Moreover, \textbf{FedEPL} consistently fails to achieve the high accuracy levels of FedEP, such as 98\% and 99\% when converges, a phenomenon also observed with ComplEx. Using ComplEx on FB15k-237-R5 and FB15k-237-R3, \textbf{FedS} surpasses \textbf{FedEPL} in MRR by 0.76\% and 2.91\%, respectively, with communication cost reductions of 7.14\% and 18.2\%. For RotatE, \textbf{FedEPL} can only attain the 98\% accuracy of \textbf{FedEP} across these three datasets and fall short of reaching the 99\% threshold. Furthermore, on the FB15k-237-R10 dataset, \textbf{FedS} demonstrates a superior MRR by 0.79\% over \textbf{FedEPL}, while simultaneously reducing communication costs by 27.78\%. Although \textbf{FedEPL} exhibits competitiveness in the metric of R@CG on FB15k-237-R5, it lags behind \textbf{FedS} due to a 0.47\% deficit in MRR when transmitting the same quantity of parameters. Moreover, \textbf{FedS} outperforms \textbf{FedEPL} by requiring 30 fewer communication rounds to achieve the convergence accuracy of \textbf{FedEPL} and show potential to attain even higher convergence accuracy.
\begin{table}[h] 
\centering
\begin{threeparttable}
\tabcolsep=0.05cm
\renewcommand\arraystretch{1.6}	
\caption{Comparison between \textbf{FedS} and \textbf{FedEPL} on three datasets. Bold values indicate the best results, while boxed values do not reach the threshold of 98\% of MRR@GC of \textbf{FedEPL}. Underlined values signify attainment of 98\% but not reaching 99\%.}
\begin{tabular}{llcc cc cc cc} 
\hline 
\multirow{2}{*}{KGE}&  \multirow{2}{*}{Setting}  & \multicolumn{2}{l}{FB15k-237-R10}&  &\multicolumn{2}{l}{FB15k-237-R5}&  &\multicolumn{2}{l}{FB15k-237-R3}\\
\cline{3-4}
\cline{6-7}
\cline{9-10}
&&MRR & R@CG & &MRR & R@CG & &MRR & R@CG\\
\cline{1-10}
\multirow{2}{*}{TransE} & FedEPL &\boxed{0.3421}  &380  &&\boxed{0.3524}  & 300 &&\boxed{0.3501}  &185\\
                  & FedS &\textbf{0.3541} &\textbf{165}  &&\textbf{0.3618}  &\textbf{105}  &&\textbf{0.3588}   &\textbf{105}\\
                  
\cline{1-10}
\multirow{2}{*}{RotatE}  & FedEPL &\underline{0.3597}  &270  &&\underline{0.3671} &\textbf{170}  &&\underline{0.3656} &\textbf{95}\\
                  & FedS &\textbf{0.3676}  &\textbf{195}  &&\textbf{0.3718}  &\textbf{170}  &&\textbf{0.3686}\tnote{*}  &120\\
                  
\cline{1-10}
\multirow{2}{*}{ComplEx}  & FedEPL &-  &-  &&\boxed{0.2953}  &70  &&\boxed{0.2879}   &55\\
                  & FedS &-  &-  &&\textbf{0.3029}  &\textbf{65}  &&\textbf{0.3170}  &\textbf{45}\\
                  
\hline 
\end{tabular}

\begin{tablenotes}
        \item[*] When \textbf{FedS} achieves MRR of 0.3656, the communication rounds are 60.
\end{tablenotes}
\label{main_result3}
\end{threeparttable}
\end{table}
% \footnotetext{}

\subsection{Ablation Study}

\begin{figure}[h]
  % \vspace{-2 em}
  \centering
    \begin{minipage}{0.49\linewidth}
		\centering
		\includegraphics[width=0.99\linewidth]{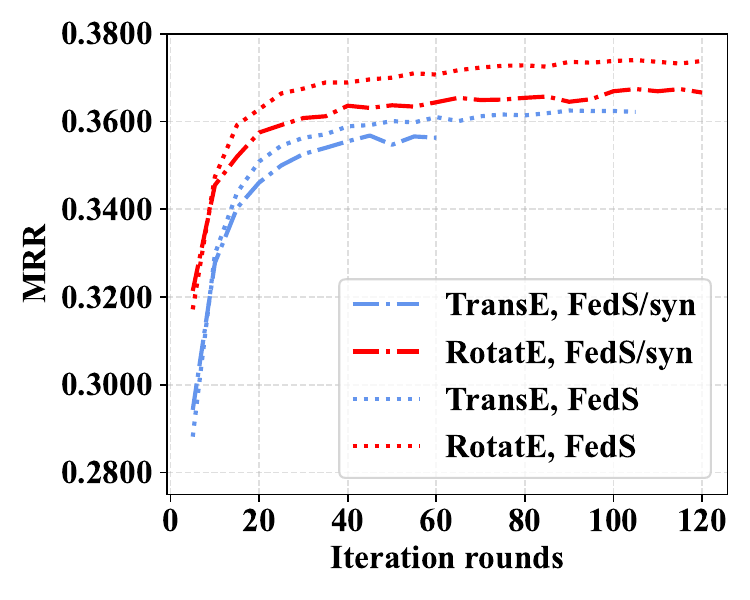}
		\subcaption{\scriptsize{FB15k-237-R3}}
		\label{a_r3}
    \end{minipage}
     \begin{minipage}{0.49\linewidth}
		\centering
		\includegraphics[width=0.99\linewidth]{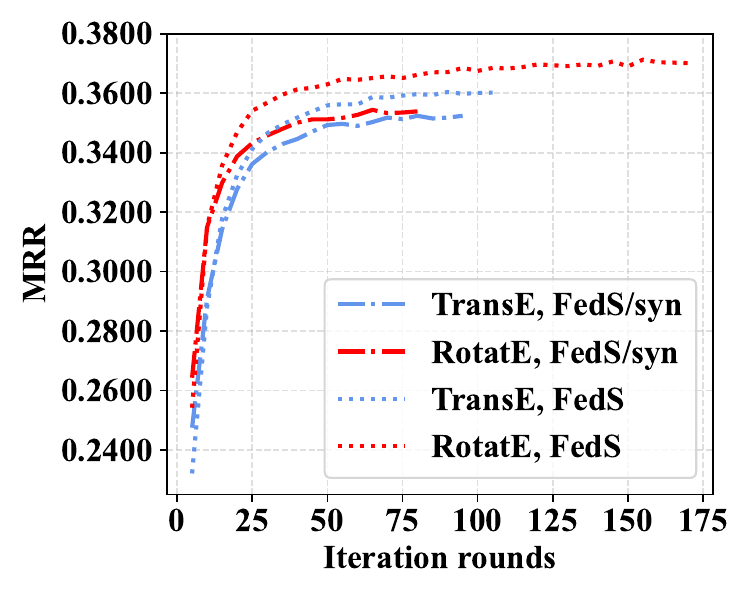}
		\subcaption{\scriptsize{FB15k-237-R5}}
		\label{a_r5}
    \end{minipage}
  \caption{Performance comparison between \textbf{FedS} and \textbf{FedS/syn}.}
  \label{a_r35}
  % \vspace{-2 em}
\end{figure}

To validate the efficacy of the Intermittent Synchronization Mechanism module, we omit it from the \textbf{FedS} (referred to as \textbf{FedS/syn} for clarity) and proceed with experiments on FB15k-237-R3 and FB15k-237-R5 utilizing the KGE methods TransE and RotatE. The corresponding outcomes are illustrated in Figure \ref{a_r3} and \ref{a_r5}.

% \begin{figure}
% \vspace{-2 em}  
% \centerline{\includegraphics[height=1.6 in]{figure/ablation_R3.pdf}}
% \caption{Results of \textbf{FedS/syn} on dataset FB15k-237-R3.} \label{a_r3} 
% \end{figure}
% \begin{figure}
% \vspace{-2 em}  
% \centerline{\includegraphics[height=1.6 in]{figure/ablation_R5.pdf}}
% \caption{Results of \textbf{FedS/syn} on dataset FB15k-237-R5.} \label{a_r5} 
% \end{figure}

The graphical representations indicate that, generally, the communication rounds required by \textbf{FedS/syn} are fewer than those of \textbf{FedS}, with the exception that RotatE is employed as the KGE method on FB15k-237-R3. Nonetheless, it is consistently observed that \textbf{FedS} achieves superior accuracy upon convergence compared to \textbf{FedS/syn}. More importantly, it is evident that as the communication rounds increase, the accuracy of \textbf{FedS} always consistently surpasses that of \textbf{FedS/syn} across all instances. These findings collectively corroborate the efficacy of the Intermittent Synchronization Mechanism module.

\subsection{Model Analysis about Various Parameters}

In this section, we study how parameters variations influence our proposed model \textbf{FedS}. We follow the same setting described in Section \ref{es}, except for the parameters under investigation.

\begin{table}[h]
\centering
\tabcolsep=0.08cm
\vspace{0.2em}
\renewcommand\arraystretch{1.4}
\caption{Comparison between \textbf{FedS} and \textbf{FedEP} across different local epochs, using TransE as the KGE Method on FB15k-237-R10.}
\begin{tabular}{cll lllll} 
\hline 
\multirow{2}{*}{Local Epoch}&  \multirow{2}{*}{Setting}  & & \multicolumn{5}{c}{Metrics}\\
\cline{4-8}

&&&MRR &Hits@10 & P@CG &P@99 &P@98 \\
\cline{1-8}
\multirow{2}{*}{2}       
                  & FedEP &&\textbf{0.3525} &\textbf{0.6080} &1.00x &1.00x &1.00x \\
                  & FedS &&0.3513 &0.6057 &0.4256 &0.4340x &0.4268x \\
                  
\cline{1-8}
\multirow{2}{*}{3}       
                  & FedEP &&0.3517 &0.6104 &1.00x &1.00x &1.00x \\
                  & FedS &&\textbf{0.3541} &\textbf{0.6121} &0.5238x &0.4411x &0.4539x \\
                  
\cline{1-8}
\multirow{2}{*}{4}       
                  & FedEP &&\textbf{0.3541} &\textbf{0.6125} &1.00x &1.00x &1.00x \\
                  & FedS &&0.3535 &0.6114 &0.4190x &0.5238x &0.4802 \\
\cline{1-8}
\multirow{2}{*}{5}       
                  & FedEP &&\textbf{0.3546} &\textbf{0.6156} &1.00x &1.00x &1.00x \\
                  & FedS &&0.3539 &\textbf{0.6156} &0.5238x &0.4889x &0.5238x \\
\hline 
\end{tabular}
\label{local_epoch}
\end{table}

We first examine the impact of varying local epochs on the performance of \textbf{FedS} using TransE as the KGE method on the FB15k237-Fed10 dataset. Four different local epochs (2, 3, 4, 5) are chosen. The result is shown in Table \ref{local_epoch}. Overall, \textbf{FedS} maintains performance levels comparable (even showing a marginal improvement in some cases) to \textbf{FedEP} across all cases, while significantly minimizing communication costs. Besides, there is no clear correlation between the effectiveness of \textbf{FedS} and the number of local epochs for most metrics, with the exception of P@98. For instance, at a local epoch of 4, the communication cost at convergence (i.e., P@CG) is lower than that observed for local epochs of 3 and 5. Although there is a slight increase in P@98 with the rise in the number of local epochs, its value remains relatively low and does not exceed the maximum of P@CG and P@99 in any scenario.

Subsequently, we investigate the influence of varying batch sizes on the performance of \textbf{FedS} using TransE as the KGE method on the FB15k237-Fed10 dataset. Three different batch sizes (128, 256, 512) are chosen. 

\begin{table}[h]
\centering
\tabcolsep=0.08cm
\vspace{0.2em}
\renewcommand\arraystretch{1.4}
\caption{Comparison between \textbf{FedS} and \textbf{FedEP} across different batch sizes, using TransE as the KGE Method on FB15k-237-R10.}
\begin{tabular}{cll lllll} 
\hline 
\multirow{2}{*}{Batch Size}&  \multirow{2}{*}{Setting}  & & \multicolumn{5}{c}{Metrics}\\
\cline{4-8}

&&&MRR &Hits@10 & P@CG &P@99 &P@98 \\
\cline{1-8}
\multirow{2}{*}{128}       
                  & FedEP &&\textbf{0.3538} &\textbf{0.6156} &1.00x &1.00x &1.00x \\
                  & FedS &&0.3519 &0.6127 &0.3175x &0.6984x &0.5238x \\
                  
\cline{1-8}
\multirow{2}{*}{256}       
                  & FedEP &&\textbf{0.3552} &\textbf{0.6148} &1.00x &1.00x &1.00x \\
                  & FedS &&0.3547 &0.6136 &0.5069x &0.4656x &0.5238x \\
                  
\cline{1-8}
\multirow{2}{*}{512}       
                  & FedEP &&0.3517 &0.6104 &1.00x &1.00x &1.00x \\
                  & FedS &&\textbf{0.3541} &\textbf{0.6121} &0.5238x &0.4411x &0.4539x \\
\hline 
\end{tabular}
\label{batch_size}
\end{table}

The result is shown in Table \ref{batch_size}. Likewise, \textbf{FedS} consistently maintains performance levels akin to those of \textbf{FedEP} across all scenarios while concurrently achieving notable reductions in communication overhead. Besides, with an increase in batch size, the gains in communication efficiency of \textbf{FedS} diminish in achieving convergence accuracy (i.e., P@CG), but escalate in attaining the other two convergence accuracies (i.e., P@99 and P@98). The prediction accuracy of the model fluctuates with the increase of batch size. For instance, when the batch size is set to 256, its MRR surpasses that of the other two cases. The determination of batch size should take into account both model performance and enhancements in model communication efficiency comprehensively.

% \begin{table*}[h] 
% \centering
% \footnotesize	
% \begin{tabular}{llll ll ll ll} 
% \hline 
% \multirow{2}{*}{KGE}&  \multirow{2}{*}{Setting}  & \multicolumn{2}{l}{FB15k-237-Fed10}&  &\multicolumn{2}{l}{FB15k-237-Fed5}&  &\multicolumn{2}{l}{FB15k-237-Fed3}\\
% \cline{3-4}
% \cline{6-7}
% \cline{9-10}
% &&MRR@CG & R@CG & &MRR & R@CG & &MRR@CG & R@CG\\
% \cline{1-10}
% \multirow{2}{*}{TransE} & FedE' &\textbf{0.3388}  &1  &&\textbf{0.3523}  &1  &&\textbf{0.3482}  &1\\
%                   & FedS &0.3541 &61.11\%  &&0.3618  &37.50\%  &&0.3588   &65.63\%\\
                  
% \cline{1-10}
% \multirow{2}{*}{RotatE}  & FedE' &\textbf{0.3581}  &1  &&\underline{0.3671} &1  &&0.3677 &1\\
%                   & FedS &0.3676  &73.58\%  &&0.3718  &121.43\%  &&0.3686  &70.59\%\\
                  
% \cline{1-10}
% \multirow{2}{*}{ComplEx}  & FedE' &-  &-  &&\textbf{0.2991}  &1  &&\textbf{0.2867}   &1\\
%                   & FedS &-  &-  &&0.3029  &92.86\%  &&0.3170  &75.00\%\\
                  
% \hline 
% \end{tabular}
% \caption{Comparison of accuracy and communication efficiency between FedS and FedE' on three datasets (FedE' is obtained from FedE by reducing embedding dimension to the same communication volume with FedS across five consecutive rounds). $MRR@CG$ means the MRR value when the model converges. $R@CG$ means communication rounds when the model converges. The value $R@CG$ of FedE' is set as 1 and that of FedS is the relative to FedE'. Values depicted in bold do not attain a threshold of 98\% in FedE, whereas those underlined signify achievement of 98\% but not attaining 99\%. }
% \label{main_result3}
% \end{table*}

\section{Conclusion}
We, by extensive experiments, first found that universal embedding precision reduction for all entities during compression greatly slows down the convergence speed, highlighting the importance to preserve embedding precision. Then, we further proposed the method \textbf{FedS}, 
based on the Entity-Wise Top-K Sparsification strategy. It can reduce the bidirectional communication overhead, and be compatible with many FKGE methods as a constituent. During upload, clients dynamically identify and transmit only the Top-K entity embeddings with the most significant changes to the server. For downloads, the server first performs personalized embedding aggregation tailored to each client. It then identifies and sends the Top-K aggregated embeddings back to each client. Additionally, \textbf{FedS} employs an Intermittent Synchronization Mechanism to alleviate the negative impact of embedding inconsistencies among shared client entities cased by federated knowledge graph heterogeneity.
Extensive experiments across three datasets validated the effectiveness of \textbf{FedS}.

\newpage
\section{Appendix}
\subsection{\textbf{FedE-KD}} \label{kd}

We applied Knowledge Distillation strategy to FKGE method \textbf{FedE} and denote it as \textbf{FedE-KD}. 

In \textbf{FedE-KD}, each client maintains both low- and high-dimensional embeddings for each entity and relation. After local training in each communication round, each client sends its low-dimension entity embeddings to server. The server conducts embedding aggregation and sends aggregated embedding to each client in the same way as \textbf{FedE}. Each client also updates their local low-dimensional entity embeddings with those from server in the same way as \textbf{FedE}. Different from \textbf{FedE}, during the local training of each client, \textbf{FedE-KD} simultaneously train the low- and high-dimensional embeddings on the supervision of local data, and also let the low- and high-dimensional embeddings distill knowledge from each other. The high-dimensional embeddings have the potential for encoding more information than low-dimension ones, which help teach the low-dimensional ones. The low-dimensional embeddings carry information from other clients by server's aggregation, which also benefit high-dimensional ones. 

We use client $c$ with triplet set $T_c$ to explain the client local training process. We use $\mathcal{L}_{L/H}(T)$ to represent the KGE loss function with negative sampling of low- (high-)dimensional embeddings, supervised by the triplet $T$ and its negative samples. For both low-dimensional embeddings, we compute their scores about triplet $T$ together with its negative samples, and concatenate the scores into a vector and normalize it using the softmax function. The normalized score vector is denoted as $\mathbf{S}_L$. Similarly, the normalized score vector of high-dimensional embedding is $\mathbf{S}_H$. Then, we use Kullback–Leiblerdivergence (KL) between $\mathbf{S}_L$ and $\mathbf{S}_H$ to let low- and high-dimensional embedding distill knowledge mutually.

Formally, the overall loss function of \textbf{FedE-KD} during local training is as follows:

\begin{equation}
\begin{split}
    \mathcal{L} = \sum_{T \in \mathcal{T}_c}  \mathcal{L}_{L}(T) + \mathcal{L}_{H}(T) + 
    \frac{\mathbf{KL}(\mathbf{S}_L, \mathbf{S}_H) + \mathbf{KL}(\mathbf{S}_H, \mathbf{S}_L)}{\mathcal{L}_{L}(T) + \mathcal{L}_{H}(T)} 
\end{split}
\end{equation}
where the third term means the co-distillation loss. Here, we choose adaptive method to progressively elevate the influence of co-distillation with the enhancement of predicted score quality (i.e., the reduction of supervised loss).

In experiments, we set the dimension of low- and high-dimension embedding as 192 and 256, respectively. The compression ratio in each communication round is $\frac{256-192}{256} = 0.25$. The other paramters follow the same setting in Section \ref{es}. 

\subsection{\textbf{FedE-SVD} (\textbf{FedE-SVD+})} \label{svd}

We applied SVD and SVD+ strategy to FKGE method \textbf{FedE} and denote them as \textbf{FedE-SVD} and \textbf{FedE-SVD+}, respectively.

In \textbf{FedE-SVD}, after local client training in each communication round, the embedding update for each entity is converted into a matrix of dimensions $\mathbb{R}^{m\times n}$($\mathrm{N}=m \times n$, $m > n$), subsequently subjected to decomposition via SVD, wherein only the top five singular values are retained. Upon receipt of the decomposed entity embedding update matrices from all clients, the server proceeds to restore them into complete entity embedding update vectors and initiates embedding update aggregation. Subsequently, the server decomposes the aggregated embedding update employing the same method as the clients. Once the clients receive and restore decomposed matrices from the server, a new round commences.

\textbf{FedE-SVD+} imposes additional constraints on loss function of \textbf{FedE} in the final epoch of each round of local training, to further improve the low-rank properties of entity update matrices, following \cite{yang2020learning}. Specifically, in the final epoch of each round of local training, \textbf{FedE-SVD+} chooses to train the entity embedding update rather than entity embedding. For each client, it first obtains embedding update for each entity (e.g embedding update $\mathbf{e}$ of entity $e$) and decomposes it via SVD ( $\mathbf{e} = \mathbf{U}diag(\textbf{s})\mathbf{V}^T $, where $\mathbf{U}\in \mathbb{R}^{m\times n}, \mathbf{s} \in \mathbb{R}^{n}, \mathbf{V}^T\in \mathbb{R}^{n\times n}$) before the last training epoch, and then conduct training on local dataset by taking $\mathbf{U}$, $\textbf{s}$, and $\mathbf{V}^T$ as training parameters and constrain $\mathbf{U}$ and $\mathbf{V}^T$ keeping orthogonality by the following regularization term $L_r$:
\begin{equation}
L_r = \alpha \frac{1}{n^2}(||\mathbf{U}^T\mathbf{U}-\mathrm{I}||^2_F + ||\mathbf{V}^T\mathbf{V}-\mathbf{I}||^2_F)
\end{equation}
where $||\cdot||_F$ is the Frobenius norm of matrix, $n$ is the rank of $\mathbf{U}$ and $\mathbf{V}^T$ and $\alpha$ is a hyper-parameter. 

After local training, \textbf{FedE-SVD+} follows the same compression process as \textbf{FedE-SVD}. 

For both \textbf{FedE-SVD} and \textbf{FedE-SVD+}, the parameter $n$ is set as 8. For \textbf{FedE-SVD+}, the parameter $\alpha$ is set as 0.05. The other parameters follow the same setting as outlined in section \ref{es}. When embedding dimension $\mathrm{N} = 256$, the embedding update matrix has dimension $\mathbb{R}^{32\times 8}$ for TransE while $\mathbb{R}^{64\times 8}$ for RotatE and ComplEx since their entity embeddings are in Complex Space. When selecting the leading five out of the eight singular values, the transmitted parameter quantity for each entity decreases to $205 = 32\times 5 + 5 + 8\times5$ for TransE and $365 = 64\times 5 + 5 + 8 \times 5$ for RotatE and ComplEx. Correspondingly, the compression ratio in each communication round is $0.1992 = \frac{256-205}{256}$ and $0.2871 = \frac{256\times 2 - 365}{256\times2}$.

\subsection{Computation of Embedding Dimension of $\textbf{FedEPL}$}\label{ced}

According to the equation \ref{equ_ace}, when sparsity ratio $p=0.7$, synchronization interval $s=4$ and embedding dimension $\mathrm{D} = 256$, there is $R^p_c = 0.7642$. The only difference between $\textbf{FedEPL}$ and $\textbf{FedEP}$ is the embedding dimension. When transmitted parameter quantity of $\textbf{FedEPL}$ equals to $R^p_c$ of $\textbf{FedEP}$ in a cycle, the embedding dimension of $\textbf{FedEPL}$ is $256 \times R^p_c = 195.64  \approx 196$. Similarly, when setting sparsity ratio $p=0.4$ and keeping others unchanged, the embedding dimension of $\textbf{FedEPL}$ is 135. For benefiting $\textbf{FedEPL}$, the embedding dimension is calculated by rounding up. Moreover, for KGE methods RotatE and ComplEx, the embedding dimension $\mathrm{D}$ in equation \ref{equ_ace} should be set to 512, reflecting the characteristics of Complex Space. However, we still utilize 256 for the same purpose. 

\bibliographystyle{plain}
\bibliography{references.bib}

\end{document}